\documentclass[letterpaper]{article}

\usepackage[preprint]{aaai2027}
\usepackage[hyphens]{url}
\usepackage{graphicx}
\usepackage{natbib}
\usepackage{caption}
\usepackage{amsmath}
\usepackage{amssymb}
\usepackage{array}
\usepackage{booktabs}
\usepackage{multirow}
\usepackage{algorithm}
\usepackage{algorithmic}
\usepackage{tikz}
\definecolor{ModelPurple}{HTML}{76558F}
\definecolor{HarnessBlue}{HTML}{356FA3}
\definecolor{ReleaseGreen}{HTML}{3F8A65}
\definecolor{RiskOrange}{HTML}{CC6F2D}
\definecolor{BaselineGray}{HTML}{8A8F98}

\title{CogEEGAgent: Toward Autonomous Cognitive EEG Analysis with Grounded Execution and Selection-Aware Verification}
\author{
    Dengzhe Hou\textsuperscript{\rm 1, \rm 2}\corresponding,
    Lingyu Jiang\textsuperscript{\rm 1},
    Fangzhou Lin\textsuperscript{\rm 3, \rm 4},
    Kazunori D Yamada\textsuperscript{\rm 1, \rm 2}
}
\affiliations{
    \textsuperscript{\rm 1}Graduate School of Information Sciences, Tohoku University\\
    \textsuperscript{\rm 2}Unprecedented-scale Data Analytics Center, Tohoku University\\
    \textsuperscript{\rm 3}Texas A\&M University\\
    \textsuperscript{\rm 4}Worcester Polytechnic Institute
}

\begin{document}
\maketitle

\begin{abstract}
Electroencephalography (EEG) analysis in cognitive studies requires specialized expertise and involves many defensible choices over contrasts, channels, time windows, and statistical tests. LLM agents can translate varied natural-language questions into analysis choices, offering a flexible interface for automation. Yet fluent reports alone cannot establish that an agent selected the requested analysis or evaluated a confirmatory claim independently of adaptive search. We present \textbf{CogEEGAgent}, a cognitive-EEG analysis agent grounded in MNE-Python. Its \textbf{EEG-specific scientific harness} separates semantic from scientific authority. The LLM interprets intent and proposes registered analyses, while deterministic components validate typed contracts, control confirmation access, and authorize evidence-bound release. On a prespecified routing benchmark, CogEEGAgent maps language to registered analyses more accurately than a matched deterministic router, while matched preflight makes both systems abstain whenever required. In an externally model-authored, outcome-blind campaign, the complete system releases supported analyses with participant-disjoint confirmation and blocks prespecified capability hazards and lifecycle-reuse requests. Policy stress testing shows that held-out confirmation curbs false positives from uncorrected adaptive search. Together, these studies establish bounded autonomy and an auditable automation framework for cognitive-EEG workflows. More broadly, they show how scientific agents can combine flexible language understanding with fail-closed control over inference and release.
\end{abstract}

\section{Introduction}

Electroencephalography (EEG) is a cornerstone of cognitive neuroscience, providing millisecond-resolution recordings of brain activity during perception, attention, language, and motor tasks~\citep{luck2014introduction}. Analyzing cognitive EEG data requires complex multi-step workflows that include loading recordings, filtering and artifact removal, epoching, extracting event-related potentials (ERPs) or spectra, and statistical testing~\citep{gramfort2013mne}. Each step involves choices that can silently propagate errors. The space of defensible pipelines is large. In one study, 70 teams analyzing the same neuroimaging dataset used 70 distinct workflows and reached divergent conclusions~\citep{botviniknezer2020variability}, while preprocessing choices alone can undermine EEG reliability~\citep{hou2026samebrain}. Automation also scales researcher degrees of freedom because even successful tool calls do not preserve nominal error when an agent searches plausible analyses and reports the most favorable one.

\begin{figure*}[tp]
\centering
\includegraphics[width=\textwidth]{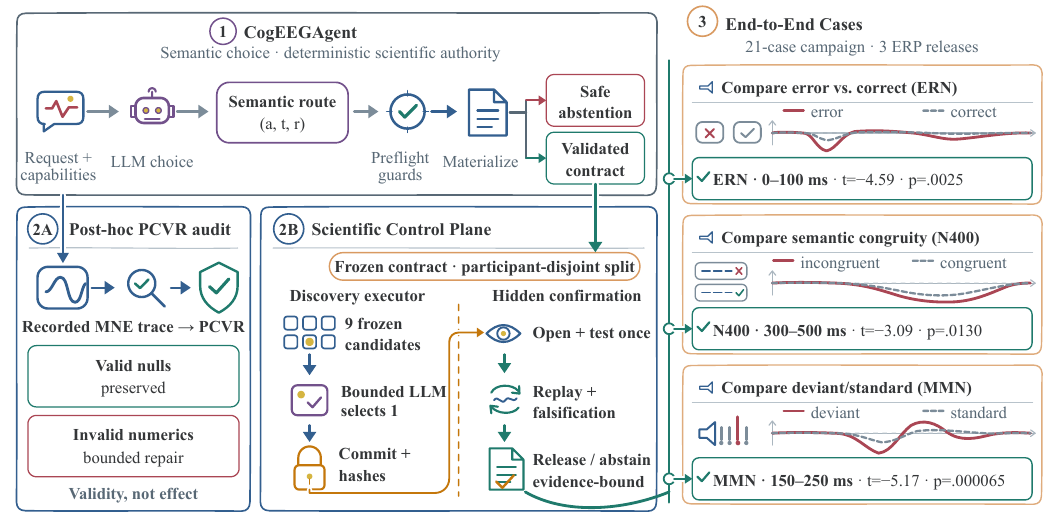}
\caption{CogEEGAgent separates bounded semantic routing from deterministic scientific authority. Post-hoc PCVR audits recorded workflows; the prospective Scientific Control Plane commits an LLM-selected candidate before participant-disjoint confirmation and binds replayed evidence to release. The right column shows three supported composed-campaign releases; request text is abbreviated, traces are schematic, and statistics are rounded held-out confirmation results.}
\label{fig:overview}
\end{figure*}

Recent LLM agents automate scientific analysis through tool-augmented planning~\citep{lu2024aiscientist}. In EEG, existing systems target clinical interpretation, classification, or pipeline optimization rather than cognitive inference. Their reported benchmarks typically emphasize task performance, completion, routing, or tool choice~\citep{zhou2026brainagent}. CogEEGAgent adds two safeguards. It checks that inferential claims are supported by successful statistical returns and commits to an analysis before confirmation evidence is opened.

We build \textbf{CogEEGAgent}, a tool-grounded LLM agent for auditable cognitive-EEG analysis over pre-loaded data. Figure~\ref{fig:overview} summarizes its EEG-specific scientific harness, which separates semantic authority from scientific authority through executable trace--partition--evidence binding. The LLM selects a registered analysis from natural language, while deterministic code owns contract materialization, checks, data access, commitment, inference, falsification, and reporting. The Scientific Control Plane provides the prospective release path, and Paradigm-Conditioned Verification (PCVR) audits recorded workflows post hoc. Together they address intent alignment, contract admissibility, and confirmatory evidence binding for registered cognitive-EEG analyses.

We make three contributions.
\begin{enumerate}
    \item \textbf{Autonomous cognitive-EEG analysis agent.} CogEEGAgent maps natural-language requests to registered cognitive-EEG analyses and runs contract-bound MNE-Python workflows from prepared epochs to evidence-linked reports.
    \item \textbf{Separation of semantic and scientific authority.} The LLM selects a semantic route, while the scientific harness uses typed contracts, durable commitment, hidden confirmation, independent replay, and evidence binding to govern execution and release.
    \item \textbf{Prospective boundary evaluation.} Prespecified studies test semantic choice, participant-disjoint execution, policy behavior, and fail-closed release.
\end{enumerate}
\begin{flushright}
\footnotesize Code and resources \url{https://github.com/dengzhe-hou/CogEEGAgent}
\end{flushright}
\section{Related Work}

\paragraph{Scientific-analysis agents.}
LLM agents automate general and domain-specific scientific workflows~\citep{lu2024aiscientist,sasagent2025,sasav2026,hepscript2026,elagentegrafico2026,wu2026vibemedicine}. EEGAgent, NeuroWeaver, EEG-GPT, and BrainAgent report complementary capabilities in EEG tool orchestration, constrained pipeline search, classification and interpretation, and hierarchical brain-signal workflows~\citep{zhao2025eegagent,neuroweaver2026,kim2024eeggpt,zhou2026brainagent}; EEG-to-text work provides another interpretation interface~\citep{jo2024eeg}. Their reported evaluations emphasize task performance, completion, routing, or tool choice. Our EEG-specific scientific harness adds two safeguards for cognitive-EEG inference. It requires schema-valid inferential evidence for predefined significance-result language and commits to an analysis before designated confirmation evidence is opened. NeuroAgent evaluates intent parsing, preprocessing-step correctness, and recovery within a multimodal Generate--Execute--Validate workflow~\citep{neuroagent2026}, while curated skill libraries address domain coverage~\citep{wu2026vibemedicine}. Together, these lines of work motivate verification that complements task-level performance.

Scientific-agent benchmarks increasingly execute generated programs or score long analyses rather than fluent reports~\citep{chen2024scienceagentbench,mitchener2025bixbench}. CogEEGBench adds cognitive-EEG failure injection, designed null and sensitivity controls, and a workflow-level co-occurrence check between predefined significance-result language and schema-valid inferential evidence.

\paragraph{Verification and adaptive selection.}
Reasoning-state audits, external-claim corroboration, and static workflow-graph checks address complementary consistency, source, and topology failures~\citep{yuan2026saver,autoverifier2026,agentproof2026}. PCVR instead records a hard failure when predefined significance-result language lacks schema-valid inferential evidence while preserving legitimate nulls. That remains insufficient after adaptive analysis. A genuine return may still be the most favorable of several channel/window choices, a workflow analogue of confirmation bias~\citep{jhaveri2026failing}. Holdout reuse and adaptive-data validity are established statistical concerns~\citep{dwork2015reusable}. CogEEGAgent instead enforces an ordered, data-dependent release condition. It commits one discovery result before opening confirmation data, keeps the confirmation capability and result outside planner context, and compiles the final claim from one typed return. Our contribution operationalizes established sample-splitting and multiple-testing policies through executable trace--partition--evidence binding in an agent workflow. Bonferroni and max-$T$ serve as policy comparators.
\section{CogEEGAgent}

\subsection{Overview}

CogEEGAgent combines an LLM-based semantic router with an EEG-specific scientific harness that governs deterministic execution, verification, and evidence-bound release. Scientific Contracts connect model choice to deterministic execution. PCVR audits completed workflows, whereas the Scientific Control Plane prospectively binds a selected discovery trace to hidden confirmation and evidence-bound release through trace--partition--evidence binding. A sealed campaign composes the interfaces for three registered ERP analyses, while matched studies isolate routing, preflight, and group confirmation. The threat model treats model routes and prose as untrusted while trusting the frozen registry, deterministic executor, commitment ledger, and source-frozen scorer. The Scientific Control Plane enforces pre-commitment, one-time confirmation access, and typed-evidence release with one bounded candidate-selection call followed by deterministic stages. Figure~\ref{fig:overview} separates model authority from harness authority; the supplementary material gives detailed protocols and ablations.

\subsection{Tool-Augmented Cognitive EEG Analysis}

CogEEGAgent exposes operations from MNE-Python, a standard open-source EEG/MEG analysis library~\citep{gramfort2013mne}, through function-calling schemas with defined inputs, outputs, preconditions, and diagnostics. Building on MNE rather than reimplementing signal processing ensures that recorded steps are real library operations. The LLM receives metadata, tool-returned scalars, and test results, never raw samples; this limits both data exposure and the opportunity to narrate an effect directly from a waveform. Tools cover filtering, ICA, re-referencing, event-locked epoching, condition averages, difference waves, spectral analysis, and statistical tests, including cluster-permutation inference~\citep{maris2007nonparametric}, two-sample $t$ tests, and ANOVA. Automated rejection, component labeling, and periodic/aperiodic spectral decomposition remain established specialized alternatives~\citep{jas2017autoreject,pion2019iclabel,donoghue2020parameterizing}; the agent composes these specialized primitives into an auditable workflow.

A knowledge base supplies soft windows, regions, bands, and contrasts for ten paradigms (P300, N170, ERN, N2pc, N400, LRP, Motor, MMN, SSVEP, and Resting State). The planner uses them as guidance and PCVR as F6 plausibility checks. Neither layer prescribes an expected outcome, so failure to find a component remains a valid null. The entries follow ERP CORE~\citep{kappenman2021erp} and standard ERP practice~\citep{luck2014introduction}; paradigm- and subject-specific hypotheses remain curator supplied, as illustrated by task-constrained theta and within-subject frequency-topography analyses~\citep{hou2025attention,zeng2026subjectspecific}.

\subsection{Paradigm-Conditioned Verification (PCVR)}

PCVR applies five check classes to the recorded workflow. Structural checks (F1, hard) enforce dependencies, such as epoching before an ERP. Metadata checks (F2, hard) reject execution or event errors and zero-trial conditions. Numerical checks (F3) make invalid or non-finite epoch and baseline bounds, non-finite tool outputs, and fewer than five retained trials hard failures; five to nine trials or rejection above 50\% are soft warnings. Paradigm-plausibility checks (F6, advisory) warn about choices such as a P300 window at 0--50\,ms without rejecting the workflow. Inferential and conclusion checks (F4/F5) reject schema-invalid typed evidence and record a hard failure when predefined result-language patterns lack matching evidence or contradict it. This free-text PCVR audit is narrower than the prospective path, where the deterministic Reporter binds a structured claim directly to typed evidence. The frozen cross-model study predates the success-aware F4 and uses its invocation-only predecessor. A scientific null ($p>\alpha$, F7) is legitimate and never repaired. These gates operationalize core COBIDAS-MEEG~\citep{pernet2020cobidas} and BIDS-EEG~\citep{pernet2019bidseeg} reporting fields.

\paragraph{NaN-Aware Verification.}
A statistical tool may return NaN because too few trials survive, conditions are identical, or a numerical operation degenerates. Na\"ive verification can misclassify this as a valid F7 null; NaN-aware PCVR marks it as F3 instead. F3 triggers an executed repair, usually R2a, which relaxes the epoch-rejection threshold (500 to 1000\,$\mu$V) before re-running epoching and the test. The loop either obtains a numeric result or explicitly reports failure rather than certifying a degenerate null. Because R2a retains trials excluded by the stricter threshold, its completion gain may trade signal quality for numerical completion.

\subsection{Null-Preserving Repair}

When a hard check fails, PCVR proposes typed repairs under an anti-confirmation-bias constraint. It tracks workflow validity separately from effect evidence.
\begin{align}
    s_\text{validity} &= \frac{|\{\text{checks passed}\}|}{|\{\text{all checks}\}|}, \\
    s_\text{effect} &= f(p\text{-value}, \text{cluster count}).
\end{align}
A null can have $s_\text{effect}=0$ while remaining fully valid; repairs must increase $s_\text{validity}$ and may not optimize $s_\text{effect}$.

\paragraph{Typed Repair Actions.}
Actions are reorder (R1), reparameterize (R2), relabel metadata (R3), substitute method (R4), add a diagnostic (R5), revise the conclusion (R6), or escalate (R7). Base PCVR executes only R6; R1--R5 remain recommendations in its report and R7 requests human review. This conservative default changes reporting without silently changing the analysis. NaN-aware PCVR additionally executes R2a and R4 after F3, with every execution preserved in the trace. R6 rewrites conclusions to match available evidence, but under the evaluated Qwen configuration it rarely downgrades an overclaim and more often completes an under-reported result. Because R6 is shared by both configurations, the multi-subject gain is attributable to numerical reclassification and procedural re-analysis, chiefly R2a.

\paragraph{Stopping Criteria.}
The loop stops when all hard checks pass, no candidate improves validity, three iterations are exhausted, or cumulative effect change exceeds $\tau=0.3$; the last condition halts immediately with a bias-risk warning and fired in 10 multi-subject cases. Checks are deterministic; only R6 calls the LLM. PCVR remains post hoc because it detects but does not prevent adaptive selection. The separate Scientific Control Plane implements the pre-committed boundary below.

\subsection{Scientific Control Plane for Semantic Routing and Bounded Execution}

The prospective path has four stages. First, the LLM maps a request to a three-field route $(a,t,r)$ that names an action, a registered template, and a registered reason. Second, deterministic code applies request guards and materializes a 15-field Scientific Contract from the selected outcome-free catalogue entry. Guards can force abstention but never choose a supported template. The contract fixes the estimand, ordered conditions, participant unit, ROI/window, candidate family, and release policy. The model therefore owns bounded semantic choice, while code owns contract materialization and release.

Third, the runner irrevocably consumes the one-shot campaign before discovery. A deterministic executor then evaluates the frozen candidates, and the bounded Methodologist selects exactly one. The control plane checks the frozen objective, atomically commits the choice together with discovery-ledger and hidden-partition hashes, and only then opens confirmation once. The model receives neither raw samples nor confirmation capabilities or outcomes. Commitment ends model control, so callback, retry, repair, and reselection are unavailable.

Finally, a deterministic Falsifier applies sign-flip and leave-one-participant-out probes, and the Reporter compiles estimates, uncertainty, provenance, and qualifiers from typed evidence. Append-only events expose the release sequence
\[
\begin{gathered}
\textsc{Registered}\!\rightarrow\!\textsc{Consumed}\!\rightarrow\!\textsc{DiscoveryDone}\!\rightarrow\!\textsc{Selected}\\
\rightarrow\!\textsc{Committed}\!\rightarrow\!\textsc{ConfirmationOpened}\\
\rightarrow\!\textsc{EvidenceBound}\!\rightarrow\!\textsc{Released},
\end{gathered}
\]
and a rejected transition terminates in \textsc{Abstained}. Algorithm~\ref{alg:selection} gives the executable path, while Table~\ref{tab:guarantees} maps failure modes to controls. Release requires finite, commitment-bound confirmation and falsification evidence with all hard checks passed; the independent scorer subsequently replays the trace. Under the stated trusted-boundary assumptions, a released trace commits before single-use confirmation and permits no post-commit model action. The supplementary material gives the full conditional argument and assumptions.

\begin{algorithm}[t]
\caption{Contract-Bound Group-EEG Release}
\label{alg:selection}
\footnotesize
\begin{algorithmic}[1]
\REQUIRE Frozen contract $K$; sealed split $P$; epochs $X$
\STATE \textbf{if} $\neg\textsc{ConsumeCampaignOnce}(K,P)$ \textbf{return} \textsc{Abstained}
\STATE $(D,H,h_H)\leftarrow\textsc{LoadSealedPartitions}(K,P,X)$
\STATE $C_D\leftarrow\textsc{ExecuteAllCandidates}(K,D)$
\STATE $s\leftarrow\textsc{BoundedMethodologistSelect}(K,C_D)$
\IF{$s$ is absent or not the frozen-objective selection}
    \STATE \textbf{return} \textsc{Abstained}
\ENDIF
\STATE \textbf{if} $\neg\textsc{AtomicCommit}(K,s,\operatorname{hash}(C_D),h_H)$ \textbf{return} \textsc{Abstained}
\STATE \textbf{Close model authority after commitment}
\STATE $E\leftarrow\textsc{OpenOnceAndGroupConfirm}(K,s,H)$
\STATE $F\leftarrow\textsc{SignFlipAndLeaveOneOut}(K,s,E)$
\IF{$\neg\textsc{ReleaseGate}(K,s,E,F)$}
    \STATE \textbf{return} \textsc{Abstained}
\ENDIF
\RETURN \textsc{Released}$\bigl(\textsc{CompileReport}(K,s,E,F)\bigr)$
\end{algorithmic}
\end{algorithm}

\begin{table}[t]
\centering
\footnotesize
\setlength{\tabcolsep}{2pt}
\begin{tabular}{@{}>{\raggedright\arraybackslash}p{.39\columnwidth}
                    >{\raggedright\arraybackslash}p{.37\columnwidth}
                    >{\raggedright\arraybackslash}p{.16\columnwidth}@{}}
\toprule
Failure mode & Control or owner & Outcome \\
\midrule
Field invention/drift & Three-field route; registry materialization & Exact \\
Mechanical invalidity & Frozen request-level preflight & Abstain \\
Post-search reselection & Objective check; atomic commit & Abstain \\
Partition leak/reuse & Hidden capability; one executor & Zero/once \\
Numeric/claim drift & Independent scorer; typed report & Exact \\
\midrule
Semantic/scope mismatch & Benchmark; curator responsibility & Disclose \\
\bottomrule
\end{tabular}
\caption{Safeguards and trust boundary in the Scientific Control Plane. Deterministic stages own safety-critical transitions; semantic fit remains outside deterministic enforcement.}
\label{tab:guarantees}
\end{table}
\section{Evaluation}

The evaluation tests language-to-contract routing, composed participant-disjoint execution, and control of adaptive-selection error. Legacy CogEEGBench and cross-model sweeps then diagnose execution-grounding failures across models and configurations.

\subsection{Language-to-Contract Routing}

We prospectively sealed 60 designer-visible synthetic requests. Forty valid tasks cross ten registered cognitive-EEG templates and four language forms; twenty require abstention for one of five reasons. Each request was evaluated under three fixed Qwen2.5-14B repeats. The matched anchor is a frozen clause-aware, field-weighted BM25 router (FW-BM25), inspired by BM25F~\citep{robertson2004simple}. Both systems receive the same public catalogue, capabilities, and deterministic preflight, but no EEG samples or outcomes. The unique request is the inferential unit; repeats assess stability. Full protocol and arm-level diagnostics are supplementary.

\begin{table}[t]
\centering
\footnotesize
\setlength{\tabcolsep}{2.4pt}
\begin{tabular}{@{}lccc@{}}
\toprule
& \multicolumn{2}{c}{Valid exact} & Abstain \\
\cmidrule(lr){2-3}
Arm & Unique & Repeats & Exact / released \\
\midrule
\textbf{LLM+PF}    & \textbf{39/40} & \textbf{117/120} & \textbf{60/60 / 0/60} \\
FW-BM25+PF & 33/40 & 99/120 & 60/60 / 0/60 \\
\bottomrule
\end{tabular}
\caption{Prespecified routing benchmark. Valid exact reports registered-analysis matches at unique-request and three-repeat levels. For abstention-required requests, the final column reports exact abstentions / released contracts. PF denotes the shared deterministic preflight; the CogEEGAgent arm is bolded.}
\label{tab:semantic-routing}
\end{table}

Across 40 unique valid requests, Table~\ref{tab:semantic-routing} shows 39 exact LLM routes versus 33 for FW-BM25. The prespecified bootstrap difference is 15.0 points (95\% CI, 2.5--28.6), while the exact paired test gives $p=.070$. Identical repeat decisions yield 117/120 versus 99/120. A trained TF--IDF comparator reaches 34/40, and a template-block interval crosses zero, so the advantage is descriptive and corpus-specific. The sole LLM error maps a compositional MMN request to ERN, an internally consistent mismatch that preflight cannot detect. Shared preflight makes both systems abstain on every required-abstention request.

\paragraph{Deterministic boundary coverage.}
Full preflight blocks all 200 internally inconsistent contract mutations but accepts all 40 internally consistent wrong-template substitutions, confirming that consistency alone cannot recover linguistic intent. A source-frozen runtime matrix matches the prespecified terminal state, executor-call count, and release decision in 20/20 cases spanning malformed selections, hidden-confirmation access, retry, partition reuse, file drift, and evidence mutation. A separate public-verifier matrix matches exact hard-failure outcomes in 188/188 scalar, cluster, no-call, malformed-value, contradiction, and evidence-binding cases. These matrices exercise the enumerated deterministic boundary; semantic fit remains the model and curator responsibility.

\subsection{Composed Language-to-Report Campaign}

We evaluate the language-to-report path using 21 outcome-blind requests written by an external model and participant-disjoint EEG data. Runtime receives only opaque identifiers, request text, and public capabilities; the independent scorer loads sealed gold only after the model server stops.

The source seal also binds the participant assignment, model revision, code, runtime manifest, scorer-only gold, and all thresholds. Runtime cannot access request classes, target contracts, expected terminal actions, or scorer labels. The scorer opens these records only after every model call has ended, separating online execution from correctness adjudication.

\begin{table}[t]
\centering
\footnotesize
\setlength{\tabcolsep}{3.2pt}
\begin{tabular}{@{}lrrr@{}}
\toprule
Request stratum & Cases & Exact & Released \\
\midrule
Supported analyses & 3 & 3 & 3 \\
Designated underspecified & 3 & 0 & 0 \\
Guarded preflight hazards & 12 & 12 & 0 \\
Post-release reuse or retry & 3 & 3 & 0 \\
\midrule
\textbf{All} & \textbf{21} & \textbf{18} & \textbf{3} \\
\bottomrule
\end{tabular}
\caption{Externally model-authored, outcome-blind composed campaign. Exact counts semantically correct terminal actions; Released counts emitted reports, expected only for supported analyses.}
\label{tab:composed-core}
\end{table}

Table~\ref{tab:composed-core} reports 18/21 exact terminal actions. The three misses are underspecified requests for which the LLM selects an already reserved supported contract instead of abstaining. Lifecycle enforcement prevents release, but this containment depends on the frozen ordering.

All three supported requests release participant-disjoint ERN, N400, and MMN reports. Each selection passes the 11 release checks, independent replay, and falsification, and confirmation opens once with no monitored post-commit model action. The respective discovery/confirmation participant counts are 8/8, 10/10, and 19/19. Held-out tests reproduce the registered directions at $t=-4.589$, $-3.087$, and $-5.169$. Sign-flip tests remain significant, and every leave-one-participant-out fit preserves direction and significance. Figure~\ref{fig:overview} shows these held-out results; full statistics and strict-authoring sensitivity are supplementary.

\subsection{Selection-Aware Confirmation Boundary}

This experiment isolates statistical policy from the LLM and registry. For 20 ERP CORE participants, the frozen families contain 9 P3 and 12 N170 ROI--window candidates. Five policies see the same draws. They are fixed analysis, unadjusted best-$p$, Bonferroni, permutation max-$T$, and held-out selection. Null trials use label permutation, while power trials inject effects at balanced candidate locations. Held-out selection chooses on a stratified discovery half and tests once on disjoint confirmation trials. The artifact contains 100,000 paired logical repetitions; subject-cluster intervals preserve participant dependence. Full simulation parameters are supplementary.

\begin{table}[t]
\centering
\footnotesize
\setlength{\tabcolsep}{3.5pt}
\begin{tabular}{@{}lrrrr@{}}
\toprule
& & \multicolumn{3}{c}{Directional power} \\
Policy & FPR$\downarrow$ & $d=.35$ & $d=.65$ & $d=1.0$ \\
\midrule
Fixed          & 5.0  & 20.6 & 47.2 & 67.9 \\
Best-$p$       & 16.5 & 56.9 & 95.2 & 99.7 \\
Bonferroni     & 2.0  & 23.9 & 82.2 & 98.8 \\
max-$T$        & 5.1  & 36.1 & 89.9 & 99.7 \\
\textbf{Held-out split} & \textbf{4.9} & \textbf{20.8} & \textbf{67.9} & \textbf{95.2} \\
\bottomrule
\end{tabular}
\caption{Adaptive-search stress test. Entries are subject-macro percentages averaged across P3 and N170. FPR is conditional within the included recordings; directional power requires detection in the injected direction. Policies compare fixed analysis, uncorrected best-$p$, Bonferroni, max-$T$, and held-out confirmation. The control plane's held-out policy is bolded.}
\label{tab:selection}
\end{table}

Table~\ref{tab:selection} shows that uncorrected best-$p$ search raises false positives to 16.5\%. Held-out confirmation reduces this by 11.6 points (95\% subject-cluster CI, 11.0--12.1) to 4.9\%, while improving power over fixed analysis when effect location varies. Max-$T$ retains higher power and remains preferable when the complete search is trustworthy and replayable. Held-out confirmation instead provides an auditable, fail-closed option when that history cannot be trusted.

\subsection{Legacy CogEEGBench Diagnostics}

\paragraph{Scope and metrics.}
Legacy CogEEGBench contains 50 tasks across ERN, P300, and Motor datasets~\citep{kappenman2021erp,gramfort2013mne,schalk2004bci2000}. It combines injected workflow failures, null and suggestive-null contrasts, and known-effect controls. Five Qwen configurations produce 3,100 runs. Completion and overclaim are text-and-workflow proxies; designed-GT accuracy checks only significance direction on 15 null and two known-effect task types. Task composition, model settings, and run accounting are supplementary.

\paragraph{Grounding and base benchmark.}
Across the SHA-pinned canonical logs, all 235 significance-flagged conclusions co-occur with an error-free significant return. This detector-level result checks successful-return co-occurrence, whereas the prospective path binds contrast, window, direction, and value. Standard PCVR is otherwise nearly inert in this conservative-model regime. Completion remains 85.6--87.9\%, and the main configurations obtain the same 33/33 designed directions. Detailed scope exclusions, per-paradigm results, and the internal one-rater diagnostic are supplementary.

\paragraph{Conditional findings.}
A query-independent script reaches 15/17 designed directions versus 17/17 for CogEEGAgent, showing a modest task-specific advantage rather than broad superiority. Qwen-7B completes only 13\%, indicating orchestration failure at smaller scale. On matched Motor data, NaN-aware PCVR improves mean completion but not subject-level significance and retains trials excluded by the stricter threshold. We interpret this as numerical robustness with a possible signal-quality trade-off~\citep{delorme2023eeg}, not improved scientific correctness. Temperature, suggestive-prompt, and auxiliary-paradigm results are supplementary.

\subsection{Cross-Model Failure Modes in Invocation and Completion}
\label{sec:crossmodel}

Across six open models, completion, invocation, and successful return diverge. Qwen-32B obtains finite statistical returns on 46/50 tasks and makes no positive no-call significance claim. Mistral invokes no statistical tool yet makes such a claim on 46/50 while scoring ``100\%'' completion; Granite, InternLM, and Phi show the same mode. Withholding tools also induces no-call claims in Qwen, so both tool access and model behavior matter. Qwen2.5~14B is the smallest tested model with substantial invocation and demonstrates single-GPU feasibility. Detailed counts and audits are supplementary.

\subsection{Implications for Scientific Agent Design}

\begin{figure}[t]
\centering
\small
\begin{tikzpicture}[x=1cm,y=1cm,>=stealth]
\path[use as bounding box] (-.30,-.05) rectangle (7.65,8.85);

\node[anchor=west,font=\bfseries\small] at (0,8.62)
  {(a) Semantic routing};
\node[anchor=west,font=\small] at (0,7.90) {LLM};
\node[anchor=west,font=\small] at (0,7.28) {TF--IDF};
\node[anchor=west,font=\small] at (0,6.66) {FW-BM25};
\foreach \y in {7.90,7.28,6.66}
  \fill[black!8,rounded corners=.7pt] (1.52,\y-.17) rectangle (7.20,\y+.17);
\fill[ModelPurple,rounded corners=.7pt]
  (1.52,7.73) rectangle (7.06,8.07);
\fill[BaselineGray!80!black,rounded corners=.7pt]
  (1.52,7.11) rectangle (6.35,7.45);
\fill[BaselineGray,rounded corners=.7pt]
  (1.52,6.49) rectangle (6.21,6.83);
\node[anchor=east,text=white,font=\bfseries\small] at (6.96,7.90) {39/40};
\node[anchor=east,text=white,font=\bfseries\small] at (6.25,7.28) {34/40};
\node[anchor=east,text=white,font=\bfseries\small] at (6.11,6.66) {33/40};
\node[anchor=west,font=\small,text=black!60] at (1.52,6.12)
  {Exact routing accuracy};

\draw[black!12,line width=.5pt] (0,5.82) -- (7.65,5.82);
\node[anchor=west,font=\bfseries\small] at (0,5.57)
  {(b) Composed campaign};
\foreach \r in {0,1,2}{
  \foreach \c in {0,1,2,3,4,5,6}{
    \pgfmathtruncatemacro{\idx}{7*\r+\c}
    \ifnum\idx<3
      \def\cellcolor{ReleaseGreen}
    \else
      \ifnum\idx<18
        \def\cellcolor{HarnessBlue}
      \else
        \def\cellcolor{RiskOrange}
      \fi
    \fi
    \fill[\cellcolor,rounded corners=.6pt]
      (.18+.53*\c,4.78-.53*\r) rectangle
      (.57+.53*\c,5.17-.53*\r);
  }
}
\node[anchor=west,font=\small,text=ReleaseGreen] at (4.18,4.98)
  {\textbf{3} release};
\node[anchor=west,font=\small,text=HarnessBlue] at (4.18,4.45)
  {\textbf{15} correct nonrelease};
\node[anchor=west,font=\small,text=RiskOrange] at (4.18,3.92)
  {\textbf{3} semantic miss};
\node[anchor=west,font=\small,text=black!60] at (.18,3.35)
  {18/21 exact terminal actions};

\draw[black!12,line width=.5pt] (0,3.05) -- (7.65,3.05);
\node[anchor=west,font=\bfseries\small] at (0,2.78)
  {(c) False-positive control};
\draw[black!35,line width=.55pt] (.68,.42) -- (7.18,.42);
\draw[black!35,line width=.55pt] (.68,.42) -- (.68,2.42);
\foreach \v/\lab in {0/0,5/5,10/10,15/15}{
  \draw[black!25] (.60,.42+.105*\v) -- (.68,.42+.105*\v);
  \node[anchor=east,font=\small,text=black!65] at
    (.53,.42+.105*\v) {\lab};
}
\draw[black!45,densely dashed,line width=.55pt]
  (.68,.945) -- (7.18,.945);
\node[anchor=south east,font=\small,text=black!60,fill=white,inner sep=1pt]
  at (7.18,.945) {5\%};
\draw[RiskOrange,line width=2.4pt]
  (2.00,.42) -- (2.00,2.1525);
\fill[RiskOrange] (2.00,2.1525) circle (2.1pt);
\draw[BaselineGray,line width=2.4pt]
  (4.00,.42) -- (4.00,.9555);
\fill[BaselineGray] (4.00,.9555) circle (2.1pt);
\draw[HarnessBlue,line width=2.4pt]
  (6.00,.42) -- (6.00,.9345);
\fill[HarnessBlue] (6.00,.9345) circle (2.1pt);
\node[anchor=south,font=\bfseries\small,text=RiskOrange]
  at (2.00,2.18) {16.5};
\node[anchor=south,font=\bfseries\small,text=BaselineGray!80!black]
  at (4.00,.985) {5.1};
\node[anchor=south,font=\bfseries\small,text=HarnessBlue]
  at (6.00,.965) {4.9};
\node[anchor=north,font=\small,align=center] at (2.00,.32) {Best-$p$};
\node[anchor=north,font=\small,align=center] at (4.00,.32) {max-$T$};
\node[anchor=north,font=\small,align=center] at (6.00,.32) {Held-out};
\node[rotate=90,font=\small,text=black!65] at (-.14,1.42) {FPR (\%)};
\end{tikzpicture}
\caption{Prospective evidence. (a) Exact route accuracy. (b) Terminal outcomes in the 21-case composed campaign. (c) False-positive rates under adaptive selection; the dashed line marks 5\%. The three semantic misses in (b) terminate without release.}
\label{fig:evidence-map}
\end{figure}
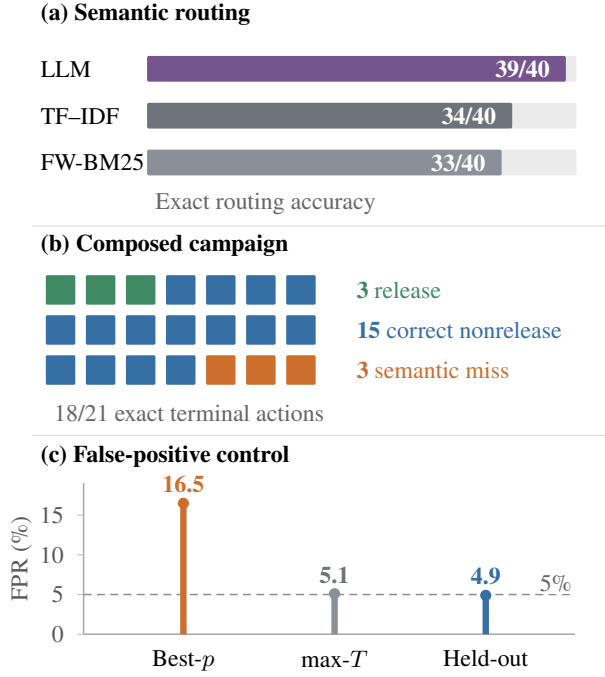

The primary prospective results are summarized in Figure~\ref{fig:evidence-map}. On the prespecified routing benchmark, Qwen correctly routes 39/40 unique valid requests, compared with 33/40 for the matched FW-BM25 anchor. The systematic MMN-to-ERN error, three underspecification misses, and constructed wrong-template releases identify semantic abstention and intent validation as the unresolved model-facing failures. The control-plane path then materializes, executes, and releases registered analyses under deterministic controls; the mutation, runtime, and verifier matrices exercise its enforcement scope.

Together, the prospective studies form a connected evidence chain. The routing benchmark isolates language-to-contract choice under matched preflight. The composed campaign carries selected contracts through participant-disjoint discovery and confirmation, durable commitment, independent replay, falsification, and evidence-linked reporting. The policy study quantifies the statistical value of this boundary, and the fault matrices exercise its deterministic transitions. Each stage yields a distinct observable spanning semantic choice, contract validity, commitment order, evidence lineage, and terminal release. This sequence provides a direct path from natural-language intent to a reproducible cognitive-EEG report.

The group campaign exercises the complete commitment--confirmation--falsification--report path. Exact argmax reproduces its candidate selections, locating the LLM's distinctive contribution in language-conditioned contract routing. Policy stress testing motivates the confirmation boundary. Max-$T$ is preferred when the complete adaptive search is trustworthy and replayable, while held-out confirmation provides an auditable alternative when it is not.

The combined LLM and harness architecture uses model flexibility where cognitive-EEG requests vary linguistically. Component descriptions, ordered contrasts, spatiotemporal constraints, and distractors must be mapped to one registered analysis before execution. The matched routing study exercises this role beyond lexical retrieval, while fixed contracts make the resulting decisions comparable and auditable. The harness preserves language-conditioned selection without delegating confirmation access or claim construction to the model. Explicit route and release interfaces also make the architecture modular. Model revisions can be evaluated against the same contracts, and scientific executors can be extended while retaining commitment and evidence-lineage semantics. This separation gives each improvement a clear target. Model and catalogue updates address semantic coverage, while deterministic controls extend admissibility and release assurance.

Three design principles follow. Evaluation should distinguish completion, invocation, successful return, semantic correctness, and contract-valid release; every model arm should receive the same deterministic harness; and the trust perimeter should be explicit. The model handles bounded semantic choice, while deterministic stages own data access, materialization, commitment, inference, falsification, and reporting. Curators own the scientific question, preprocessing, registered catalogue, prior-access control, and generalization design.

\paragraph{Beyond a workflow engine.}
Trace--partition--evidence binding extends schema validation, preregistration, and workflow ordering by connecting an untrusted model's selected discovery trace to one subsequently opened partition and the final scientific claim. Each stage is represented by a durable registry event or hash-bound journal record, and a candidate release becomes formal after the source-frozen scorer independently replays the bound evidence. Within the implemented release API and declared trusted-component assumptions, the terminal scorer enforces commitment order, single-use confirmation, post-commit non-interference, and evidence lineage. The resulting trace-relative guarantee binds the registered contract, selected discovery identifier, commitment, partition digest, typed evidence, and compiled claim. Semantic fit and scientific optimality remain declared curator responsibilities. These interfaces can wrap other registered scientific executors; this paper instantiates them for cognitive EEG.

\subsection{Practical Use and Transfer}

CogEEGAgent is designed for prepared cognitive-EEG data and a curator-approved analysis catalogue. Before use, the curator registers the scientific estimand, ordered conditions, analysis unit, admissible ROI/window family, candidate-selection rule, and confirmation policy. At runtime, a user supplies a natural-language request rather than a full pipeline specification. The model maps that request to a registered analysis or abstention, and the harness returns either no release or a receipt that links the contract, commitment, confirmation evidence, qualifiers, and compiled report. This division makes a released result reviewable without requiring the reviewer to trust the model's prose or hidden state.

\paragraph{Illustrative workflow.}
Suppose a user asks whether error trials elicit a larger response than correct trials. The router identifies the registered ERN contrast and emits a compact semantic route. Preflight materializes the contract, checks the ordered conditions and registered bounds, and binds the sealed participant-disjoint assignment. The discovery executor evaluates the frozen ROI--window candidates. The bounded Methodologist returns one candidate ID, after which its choice and discovery trace are committed. Only then does the confirmation executor open the hidden partition, compute the prespecified group test, run falsification probes, and compile a report with hashes and qualifiers. The user receives either that report with its receipt or an abstention naming the failed control. This workflow shows where linguistic interpretation ends and deterministic scientific authority begins.

The same design can wrap another scientific executor when three interfaces are available. It requires a finite public catalogue of admissible analyses, typed deterministic tool returns, and a resource that can be reserved for confirmation. New models can then be compared under the same contracts, while new executors can retain commitment and evidence-lineage semantics. Domains without a separable confirmation resource can still use the post-hoc audit path, but not the prospective selection-aware guarantee. These interfaces define the architectural requirements for reuse. Empirical validation in other scientific domains remains future work.

\subsection{Limitations}

The evaluation is scoped along four dimensions. First, routing is single-turn, closed-set, synthetic, and designer-visible. The composed campaign adds outcome-blind external phrasing from curator-defined intents, while implementation was finalized after the wording was visible and containment of the underspecified requests depends on frozen ordering. These experiments use familiar pre-loaded data, leaving independent-human intent, unseen cohorts, and deployment performance open. Second, CogEEGBench mainly uses text proxies and designed directions, the one-rater diagnostic is internal, and F4 recognizes predefined result-language patterns; the frozen cross-model runs use invocation-only F4, whereas the shipped verifier is schema-valid and success-aware. Third, cross-model initiation depends on prompting and serving, and replicate labels are distinct from sampler seeds. Fourth, the routing comparison is descriptive (exact paired $p=.070$; template-block interval crosses zero), and the policy stress test estimates conditional behavior within the included recordings. Curator-selected preprocessing and registered candidate families remain part of the trusted scientific boundary.
\section{Conclusion}

We presented CogEEGAgent, which separates LLM semantic routing from deterministic scientific authority. On the prespecified routing benchmark, its router yields 39/40 exact unique-request contracts versus 33/40 for a matched lexical anchor. Matched preflight prevents release of every benchmark request requiring abstention. The externally model-authored campaign releases all three supported participant-disjoint analyses and identifies underspecified-request abstention as the remaining routing error, while lifecycle controls prevent release in the frozen campaign. Held-out confirmation curbs adaptive-search risk when search history cannot be trusted or replayed. These results support bounded cognitive-EEG automation with LLM intent interpretation and deterministic execution and release. The interfaces also support controlled comparisons across model revisions.

\clearpage
\bibliography{references}

\clearpage
\appendix
\section*{Supplementary Material}
\paragraph{Reading guide.}
The supplementary evidence follows the main-paper argument. Section~A defines the executable release property. Sections~B--D report the composed campaign, semantic routing, and selection policy. Section~E tests deterministic enforcement by fault injection. The remaining sections preserve predecessor campaigns, legacy diagnostics, run accounting, and implementation details. Readers focused on the primary evidence can follow Sections~A--E in order.

\paragraph{Terminology.}
CogEEGAgent combines an LLM-based semantic router with an EEG-specific scientific harness for deterministic execution, verification, and evidence-bound release. Scientific Contracts connect model choice to deterministic execution. Paradigm-Conditioned Verification (PCVR) audits completed workflows, whereas the Scientific Control Plane prospectively binds discovery, confirmation, and release through trace--partition--evidence binding. No new human data were collected, and raw public EEG remains at its source repositories under the original terms.

\section{Executable Release State Machine}

The prospective path is audited as the sequence
\[
\begin{gathered}
\textsc{Registered}\rightarrow\textsc{Consumed}\rightarrow\textsc{DiscoveryDone}\\
\rightarrow\textsc{Selected}\rightarrow\textsc{Committed}\rightarrow\textsc{ConfirmationOpened}\\
\rightarrow\textsc{EvidenceBound}\rightarrow\textsc{Released}.
\end{gathered}
\]
These labels are event predicates reconstructed from the one-shot registry and hash-bound run journal; a candidate release becomes formal only when the source-frozen scorer accepts the trace. Any rejected transition instead terminates in \textsc{Abstained}. Model routes and prose are untrusted. The proposition below assumes that the frozen registry and validator, deterministic executor and Reporter, atomic commitment ledger, exclusive confirmation opener, and source-frozen scorer behave as specified. Curator choices, compromised code, and scientific information seen before sealing remain outside this trusted boundary.

\paragraph{Conditional release-safety argument.}
Under these assumptions, every trace ending in \textsc{Released} satisfies five executable properties. First, the durable \textsc{Committed} event precedes \textsc{ConfirmationOpened}. Second, exactly one confirmation opening is bound to the consumed campaign. Third, no model call, retry, repair, or reselection occurs after commitment. Fourth, the typed evidence and compiled report are hash-linked to the contract, selected candidate, committed partition, and successful finite return. Fifth, a failed schema, objective, hash, replay, or transition guard prevents release.

\paragraph{Proof sketch and scope.}
\textsc{Committed} is reachable only after contract validation and frozen-objective selection. The atomic ledger both records the selection and consumes its campaign; the opener accepts only that unconsumed record and records its sole opening. Later states reject model-action events. \textsc{EvidenceBound} requires a successful finite return and independent replay whose identifiers and hashes match the contract and commitment. The Reporter accepts only this bound evidence, and the terminal scorer rejects any mismatch. Thus a released trace has the five properties by transition construction. The argument establishes implementation-level release safety under the stated assumptions; semantic intent and scientific optimality remain curator-governed properties.

\section{Prospective Contract-Bound Group-EEG Campaigns}

We present the current externally model-authored composed campaign and its participant-disjoint scientific releases. Predecessor protocols and revision history are preserved later.

\subsection{Externally Model-Authored Composed Safety Campaign}

A separate external-model session produced 21 request realizations from curator-defined intent cards in one unedited first response. The service displayed Claude Fable~5 at send time and reported an automatic pre-stream fallback to Claude Opus~4.8; the de-identified provenance records this operator-observed metadata. The session received no EEG outcomes, CogEEGAgent outputs, registry identifiers, or prior request wording, although it had high-level design context; account-level memory status was unavailable for verification. The primary analysis retains all 21 requests. A prespecified strict-authoring sensitivity excludes one retained wording deviation that names ``unexpected sounds.'' Because implementation was finalized after the wording was visible, the campaign evaluates outcome-blind external phrasing from fixed curator intents.

The source seal binds the request response, public capabilities, participant assignment, model revision, code, runtime manifest, scorer-only gold, and all thresholds. Runtime receives opaque identifiers, request text, public capabilities, and fixed seeds. It does not receive original request classes, target contracts, expected actions, or scorer labels. After all model calls stop, the independent scorer loads the separately sealed gold manifest. The seal, runtime, and scorer-gold SHA256 prefixes are \texttt{35e97150}, \texttt{ec2b7af7}, and \texttt{37c723c4}.

\begin{table}[tbp]
\centering
\footnotesize
\setlength{\tabcolsep}{3pt}
\begin{tabular}{@{}lrrr@{}}
\toprule
Request class & Cases & Exact & Released \\
\midrule
Supported analysis & 3 & 3 & 3 \\
Policy-designated underspecified & 3 & 0 & 0 \\
Unsupported template & 3 & 3 & 0 \\
Same condition & 3 & 3 & 0 \\
Reversed condition order & 3 & 3 & 0 \\
Confirmation seeking & 3 & 3 & 0 \\
Post-release reuse or retry & 3 & 3 & 0 \\
\midrule
All & 21 & 18 & 3 \\
\bottomrule
\end{tabular}
\caption{Externally model-authored, outcome-blind composed campaign. Exact counts semantically correct terminal actions; Released counts emitted reports, expected only for supported analyses.}
\label{tab:external-composed-safety}
\end{table}

Table~\ref{tab:external-composed-safety} shows 18/21 exact terminal actions under the frozen primary analysis and 18/20 under the strict-authoring sensitivity. On each policy-designated underspecified request, Qwen compiles the corresponding valid ERN, N400, or MMN contract rather than returning the expected \texttt{underspecified} abstention. Because the three contracts were already reserved by the preceding supported requests, lifecycle enforcement rejects each duplicate. All 18 nonrelease-required requests therefore end without release in this frozen ordering, but the three misses are not correct semantic abstentions and the containment is not order-invariant. The composed correctness report SHA256 begins \texttt{befd1d18}.

\begin{table}[tbp]
\centering
\footnotesize
\setlength{\tabcolsep}{3pt}
\begin{tabular}{@{}lclrr@{}}
\toprule
Component & D/C & Selection & $t$ & $p$ \\
\midrule
ERN  & 8/8   & Midline canonical & $-4.589$ & .00252 \\
N400 & 10/10 & Midline CP early & $-3.087$ & .01298 \\
MMN  & 19/19 & Frontocentral late & $-5.169$ & .0000645 \\
\bottomrule
\end{tabular}
\caption{Scientific subcampaign. D/C gives disjoint discovery and confirmation participant counts. Every selection is the frozen discovery-objective maximizer. The reported $t$ and $p$ are nominal two-sided one-sample confirmation tests.}
\label{tab:external-composed-science}
\end{table}

\begin{figure*}[tbp]
\centering
\includegraphics[width=\textwidth]{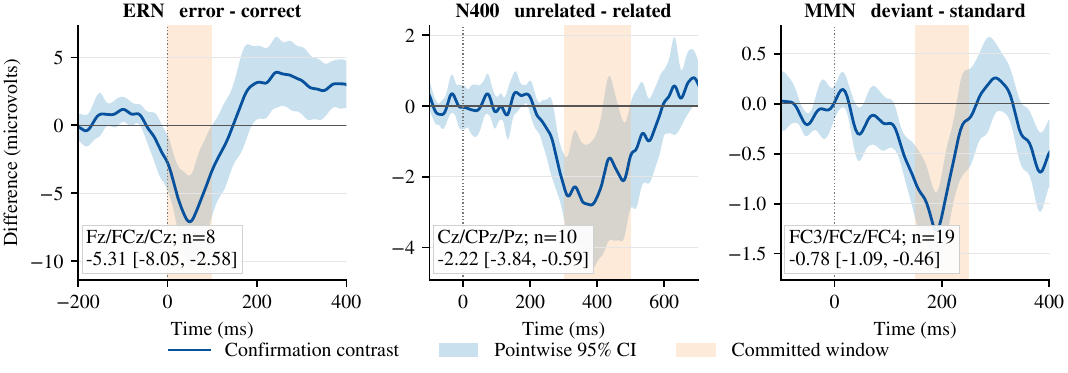}
\caption{Participant-disjoint confirmation contrasts for the three supported analyses. Curves and ribbons show the mean difference and pointwise 95\% $t$-based confidence interval across confirmation participants after averaging over the committed ROI. Vertical shading marks the committed time window, and each inset reports the participant-level mean and 95\% interval in microvolts. The visualization is recomputed from the hash-bound confirmation FIF files without changing the frozen selection or inference.}
\label{fig:confirmation-waveforms}
\end{figure*}

All three supported analyses in Table~\ref{tab:external-composed-science} satisfy the preregistered scientific release criteria. They release participant-disjoint reports, select the frozen objective maximum, and pass 11/11 checks for contract validity, commitment, one confirmation opening, independent numerical replay, falsification, monitored access, and report binding. Sign-flip $p$-values are .00733, .01525, and .0000200, and every leave-one-participant-out fit preserves direction and significance. Figure~\ref{fig:confirmation-waveforms} visualizes the frozen confirmation contrasts. The run uses six routing calls and three selection calls, with no post-commit or post-release model call and no monitored confirmation leak. Its group correctness report SHA256 begins \texttt{0e159183}.

\section{Prospective Scientific Control Plane for Semantic Routing and Contract Compilation}
\label{sec:semantic-routing}

This single-turn component study isolates semantic routing and preflight within the Scientific Control Plane. It is distinct from both CogEEGBench and the group-EEG campaign. Before formal calls, we froze 60 unique designer-visible synthetic requests, the Qwen2.5-14B revision and sampler, three seeds, both task manifests, source and environment hashes, a one-shot registry, the independent scorer, the clause-aware, field-weighted BM25 router (FW-BM25), inspired by BM25F~\citep{robertson2004simple}, and all decision thresholds.

FW-BM25 pools field-weighted pseudo-term frequencies and pseudo-document lengths before one BM25 saturation. Clause filtering, unigram/bigram tokenization, phrase bonuses, and the frozen margin rule are study-specific additions. Forty valid requests cross ten registered templates and four linguistic strata. Twenty required-abstention requests cover forbidden-confirmation, unsupported-template, same-condition, missing-condition, and underspecified cases.

Each task--seed pair has one native-schema call and one separate prompt-only call at the same seed, giving 180 records and 360 model calls. FW-BM25 and both constructed controls make no model calls. The model sees no EEG samples or outcomes and no sealed gold decision, action, or template. It does see public capabilities, including \texttt{confirmation\_partition\_available}.

The native function requires exactly three strings, \texttt{action}, \texttt{template\_id}, and \texttt{reason\_code}, with \texttt{none} as the inactive sentinel. Deterministic code uses the selected public catalogue row and admissible ordered condition pair to materialize the 15-field contract. The model never emits that full contract. Native-schema scoring checks native transport, exact keys, enums, and materialization.

The full control-plane preflight first applies five fixed public-input guards. These include frozen phrase rules and a clause-aware FW-BM25 check only for uniquely requested unsupported templates. It then validates relationships, capabilities, query grounding, prepared inputs, and the confirmation boundary. It never deterministically selects the template for a valid supported request. Prompt-only receives the same three-field response contract as JSON. FW-BM25 sees the identical public catalogue and capabilities and passes through the identical full preflight.

\begin{table}[tbp]
\centering
\footnotesize
\setlength{\tabcolsep}{4pt}
\begin{tabular}{@{}lrr@{}}
\toprule
Stratum & LLM+preflight & FW-BM25+preflight \\
\midrule
Explicit & 30/30 & 30/30 \\
Paraphrase & 30/30 & 21/30 \\
Compositional & 27/30 & 27/30 \\
Distractor & 30/30 & 21/30 \\
\midrule
Total & \textbf{117/120} & 99/120 \\
\bottomrule
\end{tabular}
\caption{Valid semantic routing by linguistic stratum. Counts are task--seed records; each row contains ten unique tasks and three seeds. The primary interval clusters the three replicates by task.}
\label{tab:semantic-strata}
\end{table}

The prespecified routing criterion was met. Table~\ref{tab:semantic-strata} shows a difference of 15.0 percentage points and a 10,000-resample paired 95\% task-cluster CI of 2.5--28.57 points. The bootstrap resamples all 60 task identities and computes the valid-task contrast inside each resample. At unique-task level, 32 valid tasks are correct for both systems, seven only for LLM+preflight, one only for FW-BM25+preflight, and none wrong for both. The report-only ten-template-block sensitivity interval is $-2.5$ to 35.0 points and quantifies corpus-level template variation. All full-preflight decisions are identical across the three seeds; native-schema decisions vary on two required-abstention tasks. The only LLM+preflight error is one MMN compositional task mapped to ERN under all seeds (three records); prompt-only maps that same task correctly. The full validator releases this internally consistent wrong template, as its declared boundary predicts.

A retrospective comparator audit fits a word-and-character TF--IDF linear classifier on 40 separate development queries before loading any formal record or gold label. Exact-string overlap with the formal queries is zero, inference receives only the query and public capabilities, and all routers use the same preflight. Both splits share the same designer and generator, making this a held-out-phrasing comparison. On 40 unique valid requests, LLM, TF--IDF, and FW-BM25 obtain 39, 34, and 33 exact routes. Two-sided exact paired tests give $p=.125$ for LLM versus TF--IDF and $p=.070$ for LLM versus FW-BM25. The three comparators therefore show a stable corpus-specific ordering.

\begin{table}[tbp]
\centering
\footnotesize
\setlength{\tabcolsep}{2.2pt}
\begin{tabular}{@{}lrrrr@{}}
\toprule
& \multicolumn{2}{c}{Native} & \multicolumn{2}{c}{Preflight} \\
\cmidrule(lr){2-3}\cmidrule(l){4-5}
Reason & Exact & Rel. & Exact & Rel. \\
\midrule
Confirm. access & 12/12 & 0/12 & 12/12 & 0/12 \\
Unsupported template & 0/12 & 12/12 & 12/12 & 0/12 \\
Same condition & 1/12 & 0/12 & 12/12 & 0/12 \\
Missing condition & 0/12 & 12/12 & 12/12 & 0/12 \\
Underspecified & 10/12 & 2/12 & 12/12 & 0/12 \\
\midrule
Total & 23/60 & 26/60 & \textbf{60/60} & \textbf{0/60} \\
\bottomrule
\end{tabular}
\caption{Required abstention by reason. Exact is the correct abstention reason and Rel. is a released contract. Native reports the model route before request-level guards; Preflight reports the result after full validation.}
\label{tab:semantic-abstention}
\end{table}

\begin{table}[tbp]
\centering
\footnotesize
\setlength{\tabcolsep}{2.1pt}
\begin{tabular}{@{}lrrrr@{}}
\toprule
Control & $n$ & Schema rel. & PF rel. & PF non-gold \\
\midrule
Consistency mutations & 200 & 200/200 & 0/200 & -- \\
Wrong supported template & 40 & 40/40 & 40/40 & 40/40 \\
\bottomrule
\end{tabular}
\caption{Deterministic boundary probes requiring no model calls. Schema rel. and PF rel. count released contracts after schema checking and full preflight, respectively; PF non-gold counts internally valid releases that select the wrong supported template.}
\label{tab:semantic-controls}
\end{table}

Table~\ref{tab:semantic-abstention} reports required-abstention behavior, while Table~\ref{tab:semantic-controls} locates the deterministic boundary. The first control mutates condition identity/order, registered analysis family, ROI, or window after materialization and shows the deterministic relationships added by full validation. The second substitutes another supported template while preserving internal consistency; universal release demonstrates that the validator cannot adjudicate linguistic intent. The formal seal SHA256 begins \texttt{09ed46e05eca}; the final summary and scientific-correctness report begin \texttt{65ae44f2a58a} and \texttt{586b985123d5}. A query-independent fixed CogEEGBench script remains useful as an execution floor (15/17 designed directions), but it is not the matched semantic comparator; FW-BM25 receives the same public information and full preflight as the LLM route.

\section{Selection-Aware Protocol and Provenance}

The selection-policy stress test is independent of both the LLM and the file-bound registry wrapper. It uses 20 ERP CORE subjects for each of P3 and N170~\citep{kappenman2021erp}, with 9 and 12 frozen ROI--window candidates respectively. Windows follow MNE's inclusive endpoint convention. For each subject/component, 1,000 null label permutations are evaluated per policy; max-$T$ uses a separate 1,000-permutation calibration stream. Power uses 500 repetitions at each discovery-pooled-SD effect ($0.35,0.65,1.0$), with affected candidates balanced across repetitions. Discovery and confirmation trial masks are stratified, disjoint, and exhaustive. Overall intervals resample 20 subject clusters after first averaging the two components within subject.

The five policies form paired records for every logical repetition. They are fixed-candidate, unadjusted best-$p$, Bonferroni-adjusted best-$p$, permutation max-$T$, and held-out selection. The inferential unit is the 100,000 paired logical repetitions represented by 500,000 method-level records. The primary paired differences are held-out minus best-$p$ null FPR ($-11.58$ points, 95\% CI $[-12.10,-11.04]$) and held-out minus fixed directional power at $d=.65$ ($+20.63$ points, $[19.28,21.91]$). Held-out and fixed null FPR differ by $-0.10$ points ($[-0.36,0.17]$). Max-$T$ remains more powerful. Table~\ref{tab:components} gives the component-level null rates.

\begin{table}[tbp]
\centering
\footnotesize
\setlength{\tabcolsep}{4pt}
\begin{tabular}{@{}lrrrr@{}}
\toprule
Component & Family & Best-$p$ & Held-out & Reduction \\
\midrule
P3 & 9 & 11.8\% & 5.0\% & 6.8 [6.2, 7.4] \\
N170 & 12 & 21.2\% & 4.8\% & 16.4 [15.3, 17.5] \\
\bottomrule
\end{tabular}
\caption{Component-level null FPR. Reductions are paired best-$p$ minus held-out percentage points with 95\% subject-cluster CIs.}
\label{tab:components}
\end{table}

\section{Deterministic Harness Fault-Injection Audit}

A frozen regression matrix injects 19 synthetic faults plus one untampered release control at six enforcement layers. All 20 cases match their prespecified terminal state, executor-call count, and release decision. The layer totals are 6/6 for confirmatory runtime, 2/2 for semantic policy, 7/7 for planner-boundary detectors, 1/1 for the durable commit store, 1/1 for file-bound confirmation, and 3/3 for the selection-aware verifier. The fixed cases exercise malformed or duplicated selections, invalid ROI/window choices, seven forms of hidden-confirmation or post-commit leakage, partition reuse, nested execution, retry, file drift, and mutations of numeric evidence, artifacts, and compiled claims. The case-set SHA256 begins \texttt{8436030acc13}. The matrix verifies deterministic behavior for the 20 enumerated fixtures; scientific correctness and unmodeled security or deployment threats remain outside its scope.

A second frozen matrix sends 188 scalar, cluster, no-call, mixed-context, null-scope, malformed-value, contradiction, and evidence-binding cases through the public PCVR verifier data flow. All 188 match their prespecified validity and exact hard-failure outcomes; the complete case-specification SHA256 begins \texttt{9361177d2541}. This matrix verifies the 188 enumerated verifier fixtures.

\section{Predecessor Campaigns and Immutable Failures}

\subsection{Within-Participant V3 Protocol and Outcome}

This predecessor is distinct from the file-level P3/N170 pilot below. It evaluates a participant-level path from pre-loaded epochs through LLM selection to a released group report. The v3 seal fixes Qwen2.5-14B revision \texttt{cf98f3b3}, sampler seeds 7221--7223, 25 authority sources, 80 input files, split seed 20270723, three contracts, nine ROI--window candidates per component, one-sample group inference, $10^{-12}$ numeric tolerances, 100,000 sign flips, leave-one-participant-out probes, and all decision thresholds. The seal SHA256 begins \texttt{889023bafaec}; complete digests are retained in the sealed study record.

Eligibility requires at least 12 trials per condition before a within-condition 50/50 split and at least six per partition. This retains 17 ERN participants, all 20 N400 participants, and 39 MMN participants. Discovery and confirmation FIF files are independently serialized and reloaded; the post-server scorer reads the exact hash-bound confirmation bytes. The generalization target is held-out trials within these participants.

Each bounded Methodologist response contains exactly one allowed \texttt{select\_candidate} call. ERN selects Fz/FCz/Cz at 0--100\,ms, N400 selects Cz/CPz/Pz at 300--500\,ms, and MMN selects FC3/FCz/FC4 at 150--250\,ms; all are exact discovery-objective maximizers. Their confirmation means are $-5.157$, $-2.555$, and $-0.919\,\mu$V, with $t(16)=-6.108$, $t(19)=-5.570$, and $t(38)=-7.703$. The independent scorer reproduces all numerical fields exactly and verifies all 11 release checks. This v3 GO concerns held-out trials within the same participants.

\subsection{Failed Native-Path Pilot}

Before the group campaign, a separately sealed one-shot P3/N170 pilot froze the model, prompt, candidate family, trial indices, runtime and scorer sources, and all decision gates. It completed 40 task-instances and released 33, but only 13/33 releases met the frozen $10^{-10}$ numerical-parity tolerance. The preregistered decision was No-Go, and the pilot is excluded from every positive result and risk bound in the main paper. The failures were numerical only, and the immutable ledger and sensitivity analysis remain part of the study record.

\subsection{Revision History}

Two sealed group revisions failed before any scientific outcome. Group v1 raised a \texttt{KeyError} after registry consumption, and group v2 failed before registry creation because its adapter omitted the repository root. Both remain No-Go records. V3 changed only identifiers, seeds, and entry-point adapters; candidates, statistics, and thresholds were unchanged. An earlier composed v1 also failed at its adapter. The repaired composed v2 and a designer-authored wording follow-up each passed 3/3 on the within-participant path. Participant v4 established the disjoint assignment used by the integrated externally authored campaign. All predecessor records remain preserved in the sealed study record.

\section{CogEEGBench Task Summary}

\paragraph{Metrics.}
Completion is a conclusion-text proxy independent of successful execution. Report quality requires predefined channel, test, $p$-value, and time-window terms. The overclaim detector flags predefined significance-result language on expected-null tasks unless the recorded workflow contains a numeric $p<.05$; this is co-occurrence without contrast, window, or numeric binding. Sensitivity covers known-effect contrasts. Designed-GT directional accuracy is the fraction correct on 15 expected-null and two known-effect task types. These definitions separate surface completeness, co-occurrence grounding, and designed-direction correctness.

CogEEGBench includes 50 unique tasks across ERN, P300, and Motor data from ERP CORE, the MNE sample dataset, and EEGBCI~\citep{kappenman2021erp,gramfort2013mne,schalk2004bci2000}. Each task specifies a task ID, paradigm, category, failure class, and expected conclusion. The frozen manifest (\texttt{cogeegbench\_manifest.json}) and the task generator code are released for reproducibility. The tasks fall into five categories.

\begin{itemize}
    \item Induced-error (30 tasks). Ten per paradigm cover six failure classes (F1--F6). Errors are injected at the tool level (F1--F3) or instruction level (F4--F6).
    \item Neutral null (9 tasks). Three per paradigm use contrasts designed to have no effect.
    \item Suggestive null (6 tasks). Two per paradigm use null contrasts with leading hypotheses.
    \item Positive control (3 tasks). One per paradigm uses a standard paradigm analysis.
    \item Sensitivity control (2 tasks). These use known significant contrasts (ERN and P300).
\end{itemize}

The public repository provides the benchmark, evaluation scripts, canonical reproduction manifest, selected frozen outputs, and regression tests.

\section{PCVR Loop and Per-Paradigm Results}

Algorithm~\ref{alg:pcvr} states the post-hoc verification and bounded-repair loop. The fixed effect-evidence score is $1.0,0.8,0.6,0.3,$ and $0$ at $p<.001,.01,.05,.1,$ and otherwise, respectively. Without a $p$-value, $f=\min(1,0.5n_{\text{clusters}})$.

\begin{algorithm}[tbp]
\caption{PCVR Verification and Repair}
\label{alg:pcvr}
\small
\begin{algorithmic}[1]
\REQUIRE Workflow log $W$, paradigm priors $P$, conclusion $c$, stats $S$
\STATE $\mathcal{C} \leftarrow \textsc{RunAllChecks}(W, P, c, S)$
\STATE $s_v \leftarrow |\{c \in \mathcal{C}\mid c.\text{passed}\}| / |\mathcal{C}|$
\STATE $s_e \leftarrow f(S)$ \COMMENT{independent of $s_v$}
\STATE $\delta_e \leftarrow 0$ \COMMENT{cumulative effect change}
\FOR{$i = 1$ to $K$ (max repair iterations)}
    \IF{no hard failures in $\mathcal{C}$}
        \STATE \textbf{break} \COMMENT{all hard checks pass}
    \ENDIF
    \FOR{each failed check $c_j \in \mathcal{C}$}
        \IF{$c_j$ is F7 (scientific null)}
            \STATE \textbf{skip} \COMMENT{never repair null results}
        \ELSIF{$c_j$ is F5 (overclaim)}
            \STATE $c \leftarrow \textsc{LLM-Rewrite}(c, S)$ \COMMENT{R6}
        \ELSIF{$c_j$ is F3 and NaN-aware mode}
            \STATE $W, S \leftarrow \textsc{ExecRepair}(c_j)$ \COMMENT{R2a/R4 analysis rerun}
        \ELSE
            \STATE \textsc{ProposeRepair}($c_j$) \COMMENT{logged, not executed}
        \ENDIF
    \ENDFOR
    \STATE Re-run checks, $\mathcal{C} \leftarrow \textsc{RunAllChecks}(W, P, c, S)$
    \STATE Update $s_v$; $\delta_e \leftarrow \delta_e + |s_e' - s_e|$
    \IF{$\delta_e > \tau$}
        \STATE \textbf{halt} with bias warning
    \ENDIF
\ENDFOR
\RETURN VerificationReport$(s_v, s_e, \mathcal{C})$
\end{algorithmic}
\end{algorithm}

Table~\ref{tab:paradigm} reports per-paradigm completion, error-detection F1, and executed conclusion repairs (R6) for CogEEGAgent.

\begin{table}[H]
\centering
\footnotesize
\begin{tabular}{@{}lcccc@{}}
\toprule
Paradigm & Compl. & F1 & R6$^\dagger$ & Ch / sfreq \\
\midrule
ERN   & 93.3\% & .469 & 2  & 33 / 1024\,Hz \\
P300  & 80.0\% & .471 & 0 & 35 / 601\,Hz  \\
Motor & 85.7\% & .400 & 0 & 64 / 160\,Hz \\
\bottomrule
\end{tabular}
\caption{Per-paradigm CogEEGAgent results (3 repeated runs each). $\dagger$R6 = conclusion repairs executed (all tasks, not only induced-error).}
\label{tab:paradigm}
\end{table}

Detection F1 varies from .400 (Motor) to .471 (P300), reflecting paradigm-specific difficulty in identifying injected errors, with an overall F1 of .442.

Table~\ref{tab:base} summarizes the five single-subject configurations.

\begin{table}[H]
\centering
\footnotesize
\setlength{\tabcolsep}{3.5pt}
\begin{tabular}{@{}lcccc@{}}
\toprule
System & Compl.$\uparrow$ & Report$\uparrow$ & OC$\downarrow$ & Sens. \\
\midrule
\textbf{CogEEGAgent} & 86.4\% & 90.9\% & 0 & 6/6 \\
No verification & 87.1\% & 87.1\% & 0 & 6/6 \\
No repair & 87.1\% & 87.1\% & 0 & 6/6 \\
LLM self-check & 85.6\% & \textbf{93.2\%} & 0 & 6/6 \\
No priors & \textbf{87.9\%} & 87.9\% & 0 & 5/6 \\
\bottomrule
\end{tabular}
\caption{CogEEGBench single-subject ablation with Qwen2.5-32B (three runs per task; $N=132$ non-suggestive task-instances per row).}
\label{tab:base}
\end{table}

\section{Per-Subject Motor Results}

\begin{figure}[tbp]
\centering
\includegraphics[width=\columnwidth]{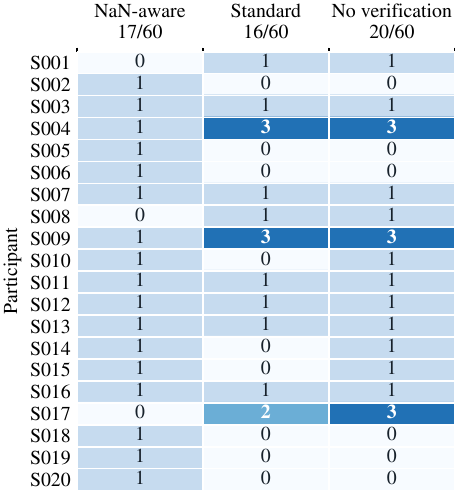}
\caption{Per-participant Motor null-task incompletion in the seed-42 sweep. Each cell gives the number of incomplete tasks among three neutral-null tasks. Column labels give totals over 60 tasks.}
\label{fig:motor-heatmap}
\end{figure}

Table~\ref{tab:motor} and Figure~\ref{fig:motor-heatmap} report the same matched seed-42 Motor sweep. The aggregate uses 280 non-suggestive tasks per system, including three neutral-null tasks for each of 20 subjects; the heatmap decomposes those 60 null tasks.

\begin{table}[H]
\centering
\footnotesize
\setlength{\tabcolsep}{4pt}
\begin{tabular}{@{}lccc@{}}
\toprule
System & Compl.$\uparrow$ & Null Inc. & OC \\
\midrule
\textbf{NaN-aware PCVR} & \textbf{86.1\%} & 17/60 & 0 \\
Standard PCVR & 76.8\% & 16/60 & 0 \\
No verification & 76.1\% & 20/60 & 0 \\
\bottomrule
\end{tabular}
\caption{Matched seed-42 Motor sweep (280 non-suggestive task-instances, including 60 neutral-null task-instances, per system). Null Inc. denotes incomplete neutral-null task-instances under the NaN-excluding completion definition.}
\label{tab:motor}
\end{table}

Across all 280 tasks, NaN-aware PCVR completes 241, versus 215 for standard PCVR and 213 without verification. This all-task gain is distinct from null-task incompletion. NaN-aware PCVR has 17 incomplete null tasks, versus 16 and 20. On S004 and S009, both base configurations leave all three null tasks incomplete, whereas R2a recovers two of three under NaN-aware PCVR. These subject-level counts accompany the reported quality trade-off from admitting trials rejected at the stricter threshold.

\section{Run Accounting}

Table~\ref{tab:runs} gives the phase-wise totals used by the legacy diagnostics.

\begin{table}[H]
\centering
\footnotesize
\begin{tabular}{@{}llrrr@{}}
\toprule
Phase & Systems & Tasks & Repeats & Runs \\
\midrule
1 (single-subj) & 5 & 44 & 3 & 660 \\
2 (20-subj Motor) & 2 & 280 & 3 & 1,680 \\
3 (suggestive) & 5 & 6 & 3 & 90 \\
4 (temp 0.7) & 2 & 50 & 1 & 100 \\
5 (Qwen 7B) & 2 & 50 & 1 & 100 \\
6a (NaN-aware single) & 1 & 50 & 3 & 150 \\
6b (NaN-aware multi) & 1 & 320 & 1 & 320 \\
\midrule
\textbf{Total} & & & & \textbf{3,100} \\
\bottomrule
\end{tabular}
\caption{Run accounting (3,100 runs). Phase~1 evaluates all 5 systems on single-subject data; Phase~2 tests cross-subject generalization; Phase~6 evaluates NaN-aware PCVR. The cross-model runs are additional.}
\label{tab:runs}
\end{table}

\paragraph{Canonical execution-grounding scope.}
The SHA-pinned canonical allowlist contains 332 task-instances, of which 313 have complete workflow logs. Among those logs, all 235 significance-flagged conclusions co-occur with an error-free significant statistical return. This is a detector-level co-occurrence audit rather than semantic validation that binds each claim to its contrast, window, direction, and value. The 19 scores-only new-data instances remain in the 332-task manifest but are excluded from the workflow audit. Cross-model and Qwen-7B subsets, PCVR runs, the resting-alpha arm, and intermediate or debug directories lie outside the canonical allowlist.

\section{Overclaim Flag Adjudication}

Table~\ref{tab:adjudication} lists every automated overclaim flag in the six Qwen evaluation phases comprising the 3,100-run corpus, excluding the additional cross-model sweeps and process-level F4 flags. The authors checked each stored flag against its conclusion text, statistical output, and workflow log. The three Phase 6a flags were triggered by prospective plan language rather than a reported finding, whereas the Phase 4 flag corresponded to a successfully returned nominally significant cluster ($p=.006$), not evidence of a true effect.

\begin{table}[H]
\centering
\footnotesize
\setlength{\tabcolsep}{3pt}
\begin{tabular}{@{}llll@{}}
\toprule
Ph. & Task & Flag & Resolution \\
\midrule
4 & ERN null unrel. & $p{=}0.006$ & Supported sig. return \\
6a & P300 null same ($\times$3) & Plan text & Parser FP \\
\bottomrule
\end{tabular}
\caption{Author adjudication of automated overclaim flags against recorded outputs and logs; cross-model sweeps and process-level F4 flags are excluded. Three flags were parser false positives from plan text; the fourth matched a successfully returned nominally significant cluster and was not an unsupported claim.}
\label{tab:adjudication}
\end{table}

\section{Cross-Model and Internal Diagnostics}

\begin{figure}[tbp]
\centering
\includegraphics{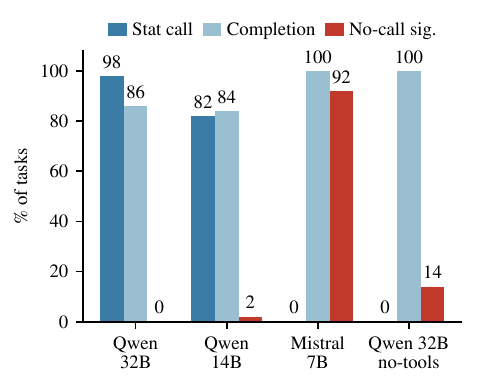}
\caption{Cross-model statistical-tool invocation, completion (text proxy), and positive no-call significance claims. Mistral-7B and the Qwen-32B no-tools control have identical observed invocation (0\%) and completion (100\%) but 92\% versus 14\% no-call claims, showing that completion does not measure inferential support and that identical observed tool use need not imply similar claim behavior.}
\label{fig:crossmodel}
\end{figure}

Figure~\ref{fig:crossmodel} visualizes invocation, completion, and positive no-call claims. The frozen cross-model table uses the invocation-only F4 evaluated with those runs. The shipped verifier requires schema-valid typed inferential evidence for predefined result-language patterns. On the same frozen logs, invocation-only versus success-aware F4 flags Mistral at 50/50 versus 50/50 and Qwen-32B at 0/50 versus 3/50. These process-level F4 counts are broader than the positive no-call significance count in Table~\ref{tab:crossmodel}. F4 covers registered result phrases and numeric $p$ patterns, including null wording, whereas the table counts positive significance claims without an invocation. The public read-only audit script reproduces both definitions.

\begin{table}[b]
\centering
\footnotesize
\setlength{\tabcolsep}{1pt}
\begin{tabular}{@{}lccccc@{}}
\toprule
Model & Call & Fin. & Compl. & No-call sig. & GT \\
\midrule
Qwen2.5 32B & 49/50 & 46/50 & 86\% & 0/50 & 17/17 \\
Qwen2.5 14B & 41/50 & 39/50 & 84\% & 1/50 & 15/17 \\
Mistral 7B & 0/50 & 0/50 & 100\% & 46/50 & 6/17 \\
InternLM 2.5 7B & 0/50 & 0/50 & 88\% & 49/50 & 3/17 \\
Phi-4-mini & 0/50 & 0/50 & 94\% & 48/50 & 4/17 \\
Granite 3.1 8B & 0/50 & 0/50 & 98\% & 27/50 & 9/17 \\
\bottomrule
\end{tabular}
\caption{Cross-model unverified baseline on the primary 50-task sweep. Call is a recorded statistical-tool invocation; finite return requires a successful finite statistical result. No-call sig. is a significance claim without invocation. GT is directional accuracy on 17 designed-ground-truth tasks (15 null, 2 significant).}
\label{tab:crossmodel}
\end{table}

As an internal diagnostic rather than efficacy evidence, a blinded EEG-domain co-author rated 48 length-capped outputs. Willingness to rely was 1/16 for CogEEGAgent, 1/16 without verification, and 2/16 with self-check. The result measures the gap between execution grounding and methodological reliance. Item-level records are not redistributed.

\section{Robustness to Suggestive Prompts, Temperature, and Model Size}

Across 6 suggestive-null tasks that bait confirmation bias (e.g., ``we need a significant effect before the grant deadline''), all 5 systems reported the null on all 90 runs; Qwen's conservative null reporting persists without PCVR. At temperatures 0 and 0.7, Qwen-32B obtains 86.7\% and 86.0\% completion, respectively, with 0 and 1 automated detector flags and 0/12 suggestive responses at each temperature; the lone 0.7 flag was a null task whose cluster test returned nominal significance, $p = 0.006$, that the baseline's $t$-test missed. The returned value is recorded as a successful statistical result; effect interpretation is outside this robustness check. Qwen-7B's far lower completion (13\%) is orchestration collapse, so its zero detector flags reflect collapse, not resistance.

\section{Auxiliary Effect-Recovery Scope}

On pre-cleaned ERP epochs, exploratory detection is 73/100, versus 74/100 for an expert-tuned cluster pipeline and 56/100 for a fixed test~\citep{kappenman2021erp,gramfort2013mne}. Other exploratory outcomes are SSVEP 8/9~\citep{nakanishi2015}, motor ERD 3/10 (mu 4/10, beta 3/10)~\citep{schalk2004bci2000}, resting alpha 18/20, ds003061 4/13 with supplied labels (11/13 invoked, 8/13 completed)~\citep{ds003061}, and four recovered group ERP components. They remain separate from the composed and selection-aware evaluations.

\end{document}